\documentclass[10pt, conference, compsocconf]{IEEEtran}
\ifCLASSINFOpdf
  \usepackage[pdftex]{graphicx}
\else
  \usepackage[dvips]{graphicx}
\fi
\usepackage{amsmath}

\usepackage{hyperref}
\usepackage{caption}
\usepackage[symbol]{footmisc}

\begin{document}
%
\title{PAI-BPR: Personalized Outfit Recommendation Scheme with Attribute-wise Interpretability}


\author{\IEEEauthorblockN{Dikshant Sagar, Sejal Bhalla, Jatin Garg, Prarthana Kansal, Rajiv Ratn Shah}
\IEEEauthorblockA{
IIIT Delhi\\
Delhi, India\\
\{dikshant17338, sejal17100, jatin17345, prarthana17357, rajivratn\}@iiitd.ac.in}
\and
\IEEEauthorblockN{Yi Yu}
\IEEEauthorblockA{
NII Japan\\
Tokyo, Japan\\
yiyu@nii.ac.jp}
}


%


\maketitle

\begin{abstract}
Fashion is an important part of human experience. Events such as interviews, meetings, marriages, etc. are often based on clothing styles. The rise in fashion industry and its effect on social influencing have made outfit compatibility a need. Thus, it necessitates an outfit compatibility model to aid people in clothing recommendation. However, due to the highly subjective nature of compatibility, it is necessary to account for personalization. Our paper devises an attribute-wise interpretable compatibility scheme with personal preference modelling which captures user-item interaction along with general item-item interaction. Our work solves the problem of interpretability in clothing matching by locating the discordant and harmonious attributes between fashion items. Extensive experiment results on IQON3000, a publicly available real-world dataset, verify the effectiveness of the proposed model. 

\end{abstract}


%
\IEEEpeerreviewmaketitle

\section{Introduction}
\label{section: intro}
\begin{figure}[!t]
    \centering
    \includegraphics[width=1\linewidth]{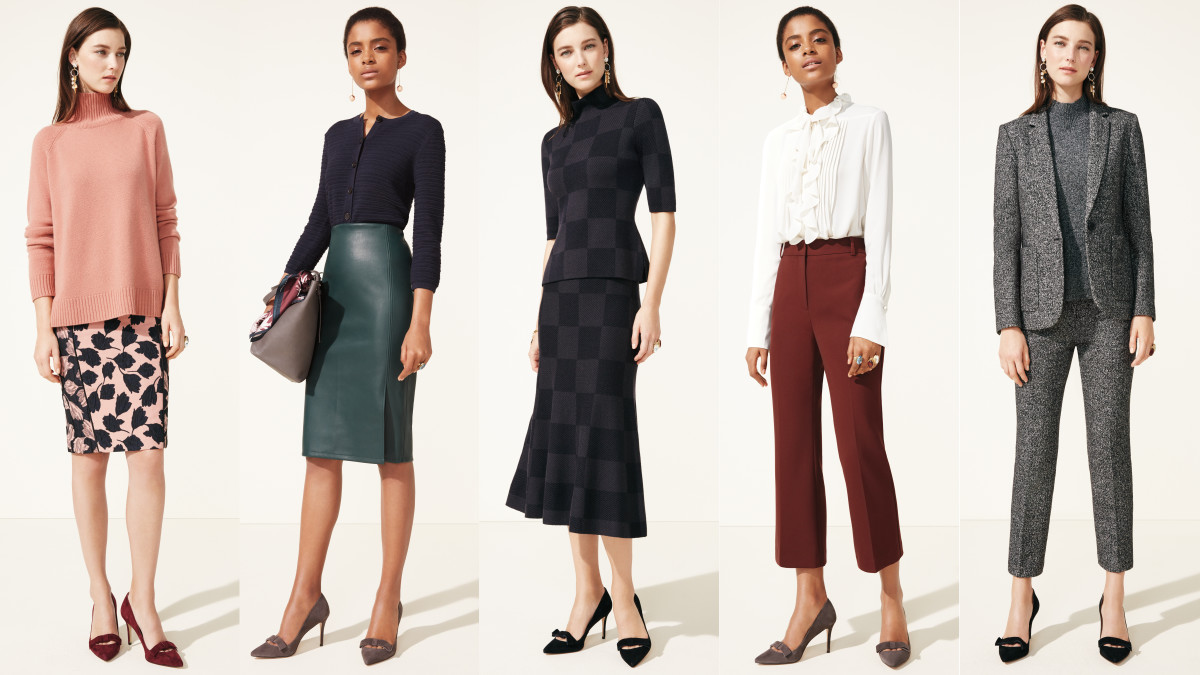}
    \caption{Example of fashionable outfits from Fashionista. (https://fashionista.com/)}
    \label{fig: exampleoutfits}
\end{figure}

The rise in fashion industry and its effect on social influencing has made outfit compatibility a need. Modern consumers choose to wear harmonious outfits (see Figure \ref{fig: exampleoutfits}), and thus, it has gained increasing research attention. However, due to the highly subjective nature of compatibility, it is desirable to account for user's personal preference in compatibility modelling. Despite its subjective nature, only a limited amount of work has been done to personalise the fashion recommendation process. For instance, most existing efforts such as the fashionability prediction \cite{li_mining_2017, simo-serra_neuroaesthetics_2015}, clothing retrieval \cite{hu_style_2014, liu_hi_2012, si_liu_street--shop_2012}, and compatibility modeling \cite{10.1145/3123266.3123394, 10.1145/3209978.3209996, 10.1145/2766462.2767755} focus on the numerical compatibility modeling between fashion items with advanced neural networks, and hence suffer from poor interpretation. Some recent works have focused on user input in the form of image or text to achieve personalization \cite{liu_hi_2012, hu_collaborative_2015}. Some have used Computer Vision algorithms to develop a real-time fashion recommendation system consistent with the latest fashion trends \cite{chao_framework_2009}. The algorithm analyzes low-level features such as shape of cloth and recommends a suitable match from existing database which is either the user's wardrobe or various websites from internet.

Fashion recommendation systems hence entail two main requirements: 1) the \textit{compatibility} of items in the recommended fashion outfits, 2) the \textit{personalization} in the recommendation process. While compatibility is a measurement of how harmonious a set of items is, personalization takes user's personal choices into account. Furthermore, in the real-world clothing matching scenarios, users may not only want to know whether the given fashion items are compatible or not, but also expect to understand the underlying reason which leads to discordance in an outfit. This leads to another significant requirement of a recommendation system, \emph{i.e.}, \textit{interpretability} of the model. The recent proliferation of online fashion communities like IQON\footnote{\href{http://www.iqon.jp/}{http://www.iqon.jp/}} or Chictopia\footnote{\href{http://www.chictopia.com/}{http://www.chictopia.com/}}, has enabled us to identify fashion choices particular to a given user. Additionally, with a suitable representation of these fashion items, it is possible to map the interactions between semantically meaningful fashion elements, hereafter referred to as \textit{attributes}. For instance, we can infer that a user prefers, say \textit{\{pear-shaped top, A-line skirt\}} over \textit{\{pear-shaped top, midi skirt\}} from her past outfit preferences (see Figure \ref{fig: bottomranking}). Apart from enhancing the recommendation, this additional information helps the user comprehend the role of attribute interactions, pinpoint the discordant attributes and harmonize the outfit accordingly.

In order to learn effective representations of fashion items, several existing works deploy Deep Learning based methods to define and measure compatibility between them \cite{song_neurostylist_2017, article}. However, the ``uninterpretability" of Deep Neural Networks has always been a limiting factor for applications requiring explanations of the learned representation \cite{fan2020interpretability}. The approach fails to explicitly map each dimension of the learned representation to the intuitive semantic aspects of fashion items. Although a few researchers have taken a step in enhancing the interpretability by modeling the outfit compatibility at an attribute-level (e.g., color and texture) \cite{feng_interpretable_2018}, their result suffers from the lack of attributes. For example, Wang et al. \cite{10.1145/3343031.3350909} limit their analysis to  low-level features (such as color and shape) and high-level features (such as category and style) without exploring them further and pinpointing the exact type of the feature (e.g., for the \textit{shape} feature, the types are \textit{hourglass-shaped}, \textit{pear-shaped}, etc). In some other works like PAICM \cite{han_prototype-guided_2019}, the issue of interpretability is addressed to a great extent but at the cost of personalization, yielding a poor performance.

\begin{figure*}[!t]
    \centering
    \includegraphics[width=1.0\linewidth]{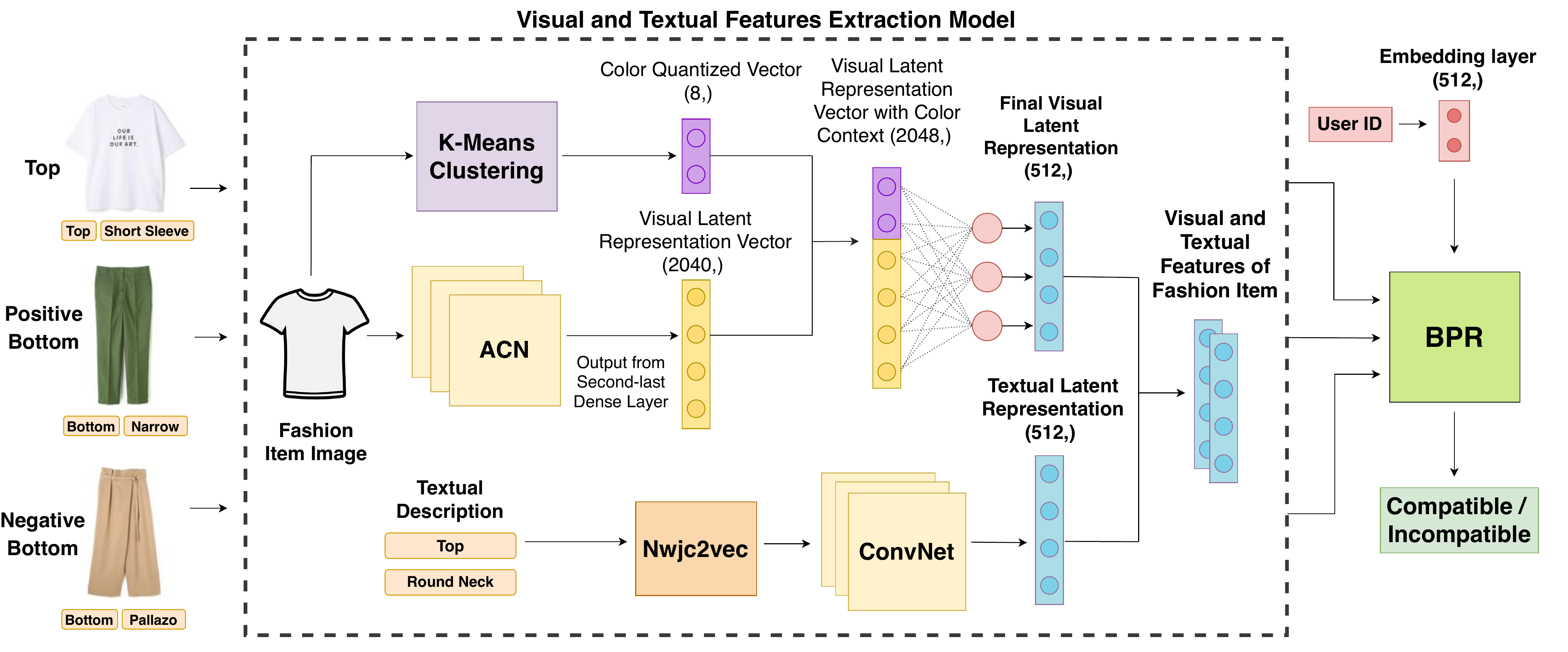}
    \caption{Feature extraction and training schematic. We obtain the semantic attribute representations for each fashion item via the pre-trained attribute classification network (ACN). The textual latent embedding is learned with Nwjc2vec followed by a Text-CNN. The visual and contextual latent embeddings are integrated with the user preferences by the BPR framework.}
    \label{fig: model}
\end{figure*}

To comprehensively tackle the aforementioned issues, we aim to devise an outfit recommendation scheme termed PAI-BPR (Personalized Attribute-wise Interpretable - BPR), as shown in Figure \ref{fig: model}, which models personalized compatibility and explains the underlying reasons for the incompatibility of a given set of items. We follow an attribute-wise interpretable scheme to explain compatibility since attributes are the most intuitive semantic cues to characterize fashion items \cite{han_prototype-guided_2019}. Without loss of generality, we focus on the problem of matching a top and a bottom. To address the same, the scheme learns the semantic attribute representation of the fashion items through deep neural networks for general compatibility modeling. To reduce the variability of the results with different datasets, the attribute classification models are pre-trained on a dataset which is richly annotated with attributes, see Section \ref{section: attributeclassification} for more details. On the other hand, the personal preference modeling focuses on extracting the latent preference factor based on the multi-modal data (textual description and image) of fashion items and hence captures the user-item interactions comprehensively. Ultimately, the model jointly regularizes the general compatibility and personal preference modeling by utilizing the Bayesian Personalized Ranking (BPR) framework \cite{rendle_bpr_2009}.

We summarise our main contributions as follows:
\begin{itemize}
    \item To the best of our knowledge, this is the first attempt to comprehensively accomplish the automatic outfit recommendation task by answering the two essential questions of compatibility determination and discordant component identification in accordance with user preference.
    
    \item Considering that both textual description and the image associated with a fashion item contribute significant information about user preferences, we build an attribute-wise interpretable personal preference modeling scheme by integrating the multi-modal data of fashion items.
    
    \item  Extensive experiments conducted on a real-world dataset verify the effectiveness of the proposed model, showing a clear advantage of using a fine-grained attribute representation for fashion item images.  
    
\end{itemize}

The remainder of this paper is structured as follows. Section \ref{section: related} briefly reviews the related work. In Section \ref{section: method}, we detail the proposed model. We present the experimental results and analyses in Section \ref{section: experiment}, followed by our concluding remarks and future work in Section \ref{section: conclusion}.

\section{Related Work}
\label{section: related}
\subsection{Fashion Analyses}
In recent years, the huge economic value of the fashion market has motivated research in the fashion domain, such as the fashion trending prediction \cite{al-halah_fashion_2017, gu_understanding_2017}, fashionability prediction \cite{li_mining_2017, simo-serra_neuroaesthetics_2015}, clothing retrieval \cite{hu_style_2014, liu_hi_2012, si_liu_street--shop_2012}, clothing parsing \cite{liang_clothes_2016}, attribute learning \cite{han_prototype-guided_2019, huang_cross-domain_2015} and compatibility modeling \cite{10.1145/3123266.3123394, 10.1145/3209978.3209996}. For example, McAuley et al. \cite{10.1145/2766462.2767755} proposed a general framework to extract visual features which model the human visual preference for a given pair of objects based on the Amazon real-world co-purchase dataset while Song et al. \cite{song_neurostylist_2017} investigated the problem of complementary fashion item matching with a multi-modal dataset collected from Polyvore\footnote{Polyvore has been acquired by the global fashion platform Ssense in 2018}. While most works focus on learning the compatibility between a pair of items, Stile \cite{jiang_who_2018} stood out as an end-to-end intelligent fashion consultant system which generates stylish outfits consisting of a top, bottom, footwear and accessories, for a given item. To further boost the performance of fashion item recommendation, Zhou et al. \cite{zhou_fashion_2018} proposed a novel approach sensitive to the current fashion trends.

\subsection{Personalized Fashion Recommendation}
A limited amount of work has been done to personalise the fashion recommendation process. Some recent works have focused on user input in the form of image or text to achieve personalization \cite{liu_hi_2012, hu_collaborative_2015}. Wang et al. \cite{wang_modeling_2014} incorporated user factor to some extent by modelling behavior of similar interest individuals and analysing retail transaction data to finally improve fashion recommendations. In the recent years, the focus of the research has shifted to explore the breadth of factors which makes a recommendation personalised. Hidayati et al. \cite{hidayati_what_2018} extracted fashion knowledge from social networking websites like Instagram to suggest items according to the body shape of a person. A similar work \cite{ma_who_2019} took features like gender, sex, body shape, etc of an individual with the occasion associated with dressing to recommend fashion items. However, the state-of-the-art system for personalised clothing matching which takes the most crucial factor of a user's personal preferences into account, is GP-BPR \cite{song_gp-bpr_2019}, which models visual and contextual features of clothing items along with personal preferences. We attempt to achieve personalization by taking inspiration from this work and supplementing it with interpretable results which increases the performance of our model while establishing the credibility of our work.

\subsection{Methodology Review}
Rising interest in the fashion domain has led to the exploration of a wide variety of implementations of  recommendation systems. Chao et al. \cite{chao_framework_2009} proposed a real-time fashion recommendation system consistent with the latest trends using computer vision algorithms. The algorithm analyzes low-level features such as shape of the cloth a person is wearing and recommends a suitable match from existing database i.e. either the user's wardrobe or various websites from internet. Adding another dimension of occasion specificity, Ma et al. \cite{ma_who_2019} used Bi-LSTM to recommend suitable outfits to the person according to various attributes of clothing, attributes of the person and the occasion. To study finer and semantic details of the cloth such as the fabric, texture etc., Al-Halah et al. \cite{al-halah_fashion_2017} used CNN representations. They predicted and forecast future trends using images and consumer data. Another CNN implementation to to mitigate the problem of two-piece clothing matching compatible with the latest fashion trends was proposed by Zhou et al. \cite{zhou_fashion_2018}. They combined the perception and reasoning models, and built two parallel CNNs to enable the system to capture the clothing features; one for the upper-body clothes and the other for the lower-body clothes. A completely new aspect of fashion recommendation was interpreted and studied by Piazza et al. \cite{piazza_emotions_2017} using Factorization Machines (FM). They investigated the impact of users' emotions on the fashion products and integrated the user's mood and emotions in the model to make the recommendation system suited to the user's preferences. Overall, most existing efforts
focus on the latent user-item interactions to tackle the
personalized recommendation problems. Beyond that, in this work, we aim to fulfil the task of personalized clothing matching by taking both user-item and item-item interactions into account, along with addressing interpretability.

\section{Methodology}
\label{section: method}
In this section, we first formally describe the research problem and its formulation; then detail the proposed model and approach (see Figure \ref{fig: model}).

\subsection{Problem Formulation}
Assume we have a set of users $ U = \{u_{1},u_{2}$, ... ,$u_{M} \} $, a set of tops $ T = \{t_{1},t_{2}$, ..., $t_{N_{t}} \} $ and a set of bottoms $ B = \{b_{1}, b_{2},$ ..., $b_{N_{b}} \} $, where $M$, $N_{t}$ and $N_{b}$ denote the total numbers of users, tops and bottoms, respectively. Each user $u_{m}$ is associated with a set of historically composed top-bottom pairs $O_{m}$ = $\{(t_{i_{1}^{m}}$, $b_{j_{1}^{m}})$, $(t_{i_{2}^{m}}$, $b_{j_{2}^{m}})$, ··· , $(t_{i_{N_{m}}^{m}}$, $b_{i_{N_{m}}^{m}})\} $, where $i_{k}^{m} \in [1, 2$, ··· , $N_{t}]$ and $j_{k}^{m} \in [1, 2,$ ··· , $N_{b}]$ refer to the index of the top and bottom.  Further, each item $t_{i} (b_{j})$ is associated with an image having a clear background, a textual description and structured category labels. In this work, we characterize each fashion item with a set of attributes (e.g., the color and category) $A$ = $\{a_{q}\}^{Q}_{q=1}$, where $a_{q}$ is the $q$-th attribute and $Q$ is the total number of attributes. Each attribute $a_{q}$ is associated with a set of elements representing its possible values $E_{q}$ = $\{e^{1}_{q}$, $e^{2}_{q}$,···, $e^{M_{q}}_{q}\}$, where $e_{q}^{i}$ refers to the $i$-th element and $M_{q}$ is the total number of elements corresponding to $a_{q}$. For simplicity, all $E_{q}$ are combined in order and hence a unified set of attribute elements $E = \cup_{q=1}^{Q}E_{q}$ = $\{e_{1}, e_{2}$,···, $e_{M}\}$ is derived, where $M = \Sigma^{Q}_{q=1} M_{q}$. 

In this work, we aim to tackle the essential task of modeling interpretable compatibility between fashion items for outfit recommendation by taking the user factor into account. Without loss
of generality, we particularly investigate the following problem, “which bottom would be preferred by the user to match the given top". Let $p^{m}_{ij}$ denotes the preference of the user $u_{m}$ towards the bottom $b_{j}$ for the top $t_{i}$. Based on this metric, we can generate a personalized ranking list of bottoms $b_{j}$’s for a given top $t_{i}$ and hence solve the practical problem
of personalized clothing matching. This ranking list is further used to infer compatible and incompatible pairs of attributes which can help us understand the underlying reason for discordance in an outfit.  

In order to accurately measure $p^{m}_{ij}$, we devise a personalized compatibility modeling network $F$, which is capable of incorporating the user's personal preferences into the compatibility modeling between fashion items as follows, 
\begin{equation}
     p^{m}_{ij} = F(t_{i}, b_{j}, u_{m} | \Theta_{F})
\end{equation}

where $\Theta_{F}$ refers to the model parameters which are to be learned.

\subsection{Semantic Attribute Learning}
Attributes play a pivotal role in both characterizing fashion items and interpreting the matching results. However, most of the existing benchmark datasets pertaining to outfit compatibility or recommendation lack attribute ground truth for fashion items. The task of acquiring accurate fine-grained attribute representations for benchmark datasets poses a challenge for us. We tackle this by deploying a modified Alexnet \cite{10.5555/2999134.2999257} architecture for each attribute learning task. While we argue that well pre-trained attribute classification networks is the most straightforward way to obtain the interpretable representations for a fashion item, we should not neglect the category information as an essential attribute as well. Therefore, after passing the image I of a fashion item into separate classification networks for each attribute (h$_{q}$ for q$^{th}$ attribute), we obtain the final semantic representation as follows:  \\* \begin{equation} f_{i} = [f_{i}^{ 1}; f_{i}^{ 2};...;f_{i}^{  Q-1}; f_{i}^{ c}] \end{equation} \\* where f$_{i}^{ q}$ = h$_{q}$(I$|\theta_{q}$), $\theta_{q}$ being the network parameter of h$_{q}$, and f$_{i}^{c}$ stands for the one-hot semantic attribute representation derived from the category labels for the i$^{th}$ item.

\begin{table}[]
\centering
\caption{Attribute categories and examples of corresponding attribute elements in DeepFashion \cite{liu_deepfashion_2016}.}
\normalsize
\begin{tabular}{|c|c|}
\hline
\textbf{Attributes}     & \textbf{Attribute Elements}            \\ \hline
Texture of clothes      & checked, striped, marble print \\ \hline
Style of clothes        & athletic, bold, biker, chic            \\ \hline
Fabric of clothes       & classic cotton, netted, woven          \\ \hline
Shape of clothes        & a-line, oversized, swing               \\ \hline
Part details of clothes & back cutout, bell-sleeve, pocket       \\ \hline
\end{tabular}
\end{table}

\subsection{PAI-BPR} 
To achieve personalized clothing matching, it is essential to account for both the item-item compatibility and the user-item preference. Hence, we measure the user preference towards a bottom for a given top by modeling both the general compatibility and the personal preferences of the user. Formally, we have,

\begin{gather}
    p^{m}_{ij} = \mu \;.\;s_{ij} + (1 - \mu)\;.\;c_{mj} \\
     s_{ij} = G(t_{i}, b_{j} | \Theta_{G}) \\
      c_{mj}  = P(u_{m}, b_{j}| \Theta_{P})
    \end{gather}

where $G$ and $P$ correspond to he general compatibility modeling and personal preference modeling networks, respectively with $\Theta_{G}$ and $\Theta_{P}$
as the corresponding model parameters. Likewise, $s_{ij}$ denotes the general compatibility between the top $t_{i}$ and bottom $b_{j}$, and $c_{mj}$ represents the personal preference of user $u_{m}$ towards the bottom $b_{j}$. $\mu$ is a non-negative tradeoff parameter to control the relative importance of both components.

The general item-item compatibility between fashion items entails complex interactions between attributes. These non-linear interactions are learned by a multi-layer perceptron (MLP) which performs excellently in various representation learning tasks \cite{wang_incremental_2018, chen_deeplab_2017, liu_cross-modal_2018}. We also leverage both the textual and visual modalities in the form of item description and image respectively, which coherently characterize a fashion item. On the other hand, for the personal preference modeling of a bottom, we resort to the matrix factorization framework, which has shown great success in personalized recommendation tasks \cite{bobadilla_recommender_2018, kim_convolutional_2016, packer_visually-aware_2018, koren_collaborative_2009}. Inspired by Song et al. \cite{song_gp-bpr_2019}, the matrix factorisation method is extended to incorporate the latent content-based preference factors. The underlying philosophy is that the user-item interaction matrix can be decomposed into the latent user and item factors, whose inner product encodes the interaction scores between a user and an item. Further, since a user's preference for an item is highly affected by the visual characteristics and the  contextual features, like the brand and material, it is important to consider the content-based factors as well.

In order to model the implicit interaction
among users and fashion items (i.e., tops and bottoms), the BPR framework is adopted since it has proven to be powerful in the pairwise preference modeling \cite{cao_embedding_2017, LILY-c160, loni_bayesian_2016}. A training set, $D$ is constructed in the following manner to train the BPR:
\begin{center}
    $D$ := $\{(m, i, j, k)|u_{m} \in U \cap (t_{i}, b_{j}) \in O_{m} \cap b_{k} \in B \setminus b_{j}\}$,
\end{center} where the quadruplet $(m, i, j, k)$ indicates that the user $u_{m}$ prefers bottom $b_{j}$ over $b_{k}$ to match the given top $t_{i}$.

\begin{table}[!ht]
\centering
\caption{Performance of attribute representation learning in terms of Area under the ROC Curve.}
\label{AttrClassification}
\normalsize
\begin{tabular}{|c|c|c|}
\hline
\textbf{Attributes}          & \textbf{Tops} & \textbf{Bottoms} \\ \hline
Texture of clothes             & 0.77        & 0.75            \\ \hline
Style of clothes            & 0.81        & 0.77            \\ \hline
Fabric of clothes           & 0.86        & 0.82            \\ \hline
Shape of clothes  & 0.83        & 0.70            \\ \hline
Part Details of  clothes           & 0.84        & 0.75           \\ \hline

\end{tabular}
\end{table}

\section{Experiment}
\label{section: experiment}
To evaluate the proposed method, we conducted extensive experiments on the real-world dataset IQON3000 \cite{song_gp-bpr_2019} by answering the following research questions:
\begin{itemize}
    \item Does the proposed model outperform the state-of-the-art methods?

    \item How does the proposed model perform in the application of the personalized complementary fashion item retrieval?
    
    \item What are various approaches to understand the predictions and establish the interpretability of our proposed model?
    
\end{itemize}

\subsection{Experimental Settings}
\subsubsection{Dataset}
To evaluate our model, we adopted the real-world dataset IQON3000 \cite{song_gp-bpr_2019} consisting of 217,806 unique outfits created by 3,568 users. Each fashion item is associated with a visual image, relevant categories and the description. Each of these images was resized to a smaller vector of size (100, 100, 3), in order to make the training process computationally efficient. All the results reported in Table \ref{table: modelcomparsion} were evaluated using these resized images. Additionally, to train the attribute classification networks and obtain the semantic attribute representations of fashion items, we utilized an auxiliary benchmark dataset DeepFashion \cite{liu_deepfashion_2016}, comprising of 299,221 fashion items, each of which is labeled by 5 attributes with a total of 1,000 attribute elements.

\subsubsection{Attribute Representation Learning}
\label{section: attributeclassification}
For learning the semantic attribute representation of fashion items, a modified AlexNet \cite{10.5555/2999134.2999257} architecture with an additional Dense Layer of size = 2040 before the final fully connected layer, was used. The output of this additional dense layer (size = (2040, 1)) was extracted and used as a representative vector of the same size. Apart from that, since the auxiliary dataset lacked the annotations for color, we quantized the image to a color vector of size (8, 1) using K-Means Clustering. This color vector was appended to the the vector obtained from the neural network to have a resultant attribute representation vector of size (2048, 1). The dataset was randomly split into training set (80\%) and testing set (20\%), and the 'binary cross-entropy' loss was chosen due to multi-label hot encoded target type to train the Attribute Classification Network. The area under the ROC curve (AUC) was adopted as the metric to evaluate the performance of the attribute representation learning. Table \ref{AttrClassification} shows the performance of several attribute examples and the corresponding attribute elements.

\begin{figure*}[!t]
    \centering
    \includegraphics[width=1\linewidth]{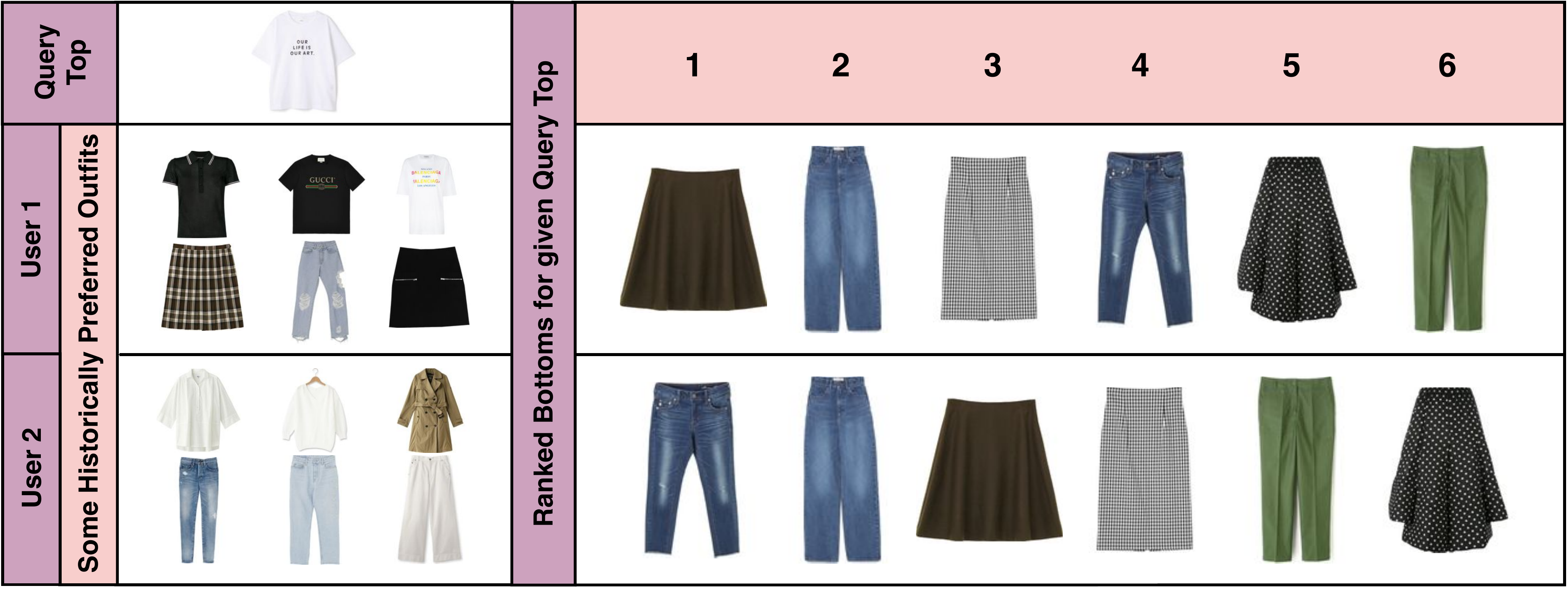}
    \caption{Illustration of the ranking results of a given set of bottoms for a given top for two different users with varying historical preferences.}
    \label{fig: bottomranking}
\end{figure*}

\subsubsection{Contextual Representation}
In this work, we consider the category metadata and title description as the contextual information of the fashion item. First of all, we used the Japanese morphological analyzer Kuromoji\footnote{\href{http://www.atilika.org/}{http://www.atilika.org/}} to tokenize the extracted text. Further, due to the the compelling success of the CNNs in various natural language processing tasks \cite{hu-etal-2016-harnessing, kim-2014-convolutional, 10.1145/2766462.2767830}, we choose them for obtaining effective contextual representations, instead of the traditional linguistic features \cite{10.1145/2766462.2767726, 10.5555/2832415.2832578}. Each word in the contextual description was represented in the form of 300-dimensional vector provided by the Japanese Word2vec called Nwjc2vec, in the search mode which is created from NINJAL Web Japanese Corpus \cite{shinnou_nwjc2vec_2017}. Overall, each contextual description was first represented as a concatenated word vector, where each row represents one constituent word. Next, a single channel CNN consisting of a convolutional layer on top of the concatenated word vectors and a max pooling layer, was deployed. More precisely, we used four kernels with sizes of 2, 3, 4, and 5, respectively and for each kernel, we had 100 feature maps. After adding rectified linear unit (ReLU) as the activation function followed by another dense layer, we ultimately obtained a 512-D contextual representation vector for each fashion item.

\begin{table}[]
\centering
\caption{Performance comparison among different approaches in terms of Area Under the ROC Curve.}
\label{table: modelcomparsion}
\normalsize
\begin{tabular}{|c|c|c|}
\hline
\textbf{Baselines/Proposed}           & \textbf{Approach}                   & \textbf{AUC}             \\ 
\hline
            & POP-T                                      & 0.6042          \\
            & POP-U                                      & 0.5951          \\
            & RAND                                       & 0.5014          \\
            & Bi-LSTM \cite{han_learning_2017}           & 0.6611          \\
 Baselines  & BPR-DAE \cite{song_neurostylist_2017}      &0.6912           \\
            & BPR-MF \cite{cao_embedding_2017}           & 0.7867          \\
            & VBPR \cite{he_vbpr_2015}                   & 0.8088          \\
            & TBPR \cite{song_gp-bpr_2019}               & 0.8102          \\
            & VTBPR \cite{song_gp-bpr_2019}              & 0.8194          \\
            & GP-BPR \cite{song_gp-bpr_2019}             & 0.8321          \\ 
\hline
            &PAI-BPR-V                                   & 0.8413 \\
Proposed    &PAI-BPR-T                                   & 0.8432 \\
            & $\quad$ \textbf{PAI-BPR} $\quad$ & $\quad$ \textbf{0.8502}$\quad$  \\ 
\hline
\end{tabular}
\end{table}

\subsection{Model Comparison}
To evaluate the proposed model, the following state-of-the-art methods were chosen as baselines.
\begin{itemize}
    \item \textbf{POP-T:} The “popularity” of the bottom was used as a measure of its compatibility with a top. It is defined as the number of outfits that the bottom appeared in the training set.
    \item \textbf{POP-U:} For this baseline, the popularity of the bottom was defined as the number of users who used this bottom as a part of an outfit in the training set.
    
    \item \textbf{RAND:} The compatibility scpres between positive and negative pairs were assigned randomly.
    
    \item \textbf{Bi-LSTM:} The bidirectional LSTM method which sequentially models the outfit compatibility by predicting the next item conditioned on the previous ones \cite{han_learning_2017}, was modified to deal with a pair of top and bottom.
    
    \item \textbf{BPR-DAE:} In this baseline the content-based neural scheme introduced by Song et al. \cite{song_neurostylist_2017}, that uses a dual autoencoder network to jointly model the coherent relation between different modalities of fashion items and the implicit preference among items was used.
    
    \item \textbf{BPR-MF:} This baseline captures the latent user-item relations by the MF method in the pairwise ranking method introduced by Rendle et al. \cite{cao_embedding_2017}
    
    \item \textbf{VBPR:} The VBPR proposed by McAuley et al. \cite{he_vbpr_2015} was adopted to exploit the visual data of fashion items with the factorization method to recommend an item for the user. 
    
    \item \textbf{TBPR:} VBPR was modified by replacing the visual signals with the textual description of fashion items \cite{song_gp-bpr_2019}. 
    
    \item \textbf{VTBPR:} This is an extended version of VBPR which includes  context factor to comprehensively characterize the user’s preference from both the visual and contextual perspectives \cite{song_gp-bpr_2019}.
    
    \item \textbf{GP-BPR:} The current state-of-the-art in personalized compatibility modeling for outfit recommendation \cite{song_gp-bpr_2019}.

\end{itemize}

\begin{figure*}[!t]
    \centering
    \includegraphics[width=1\linewidth]{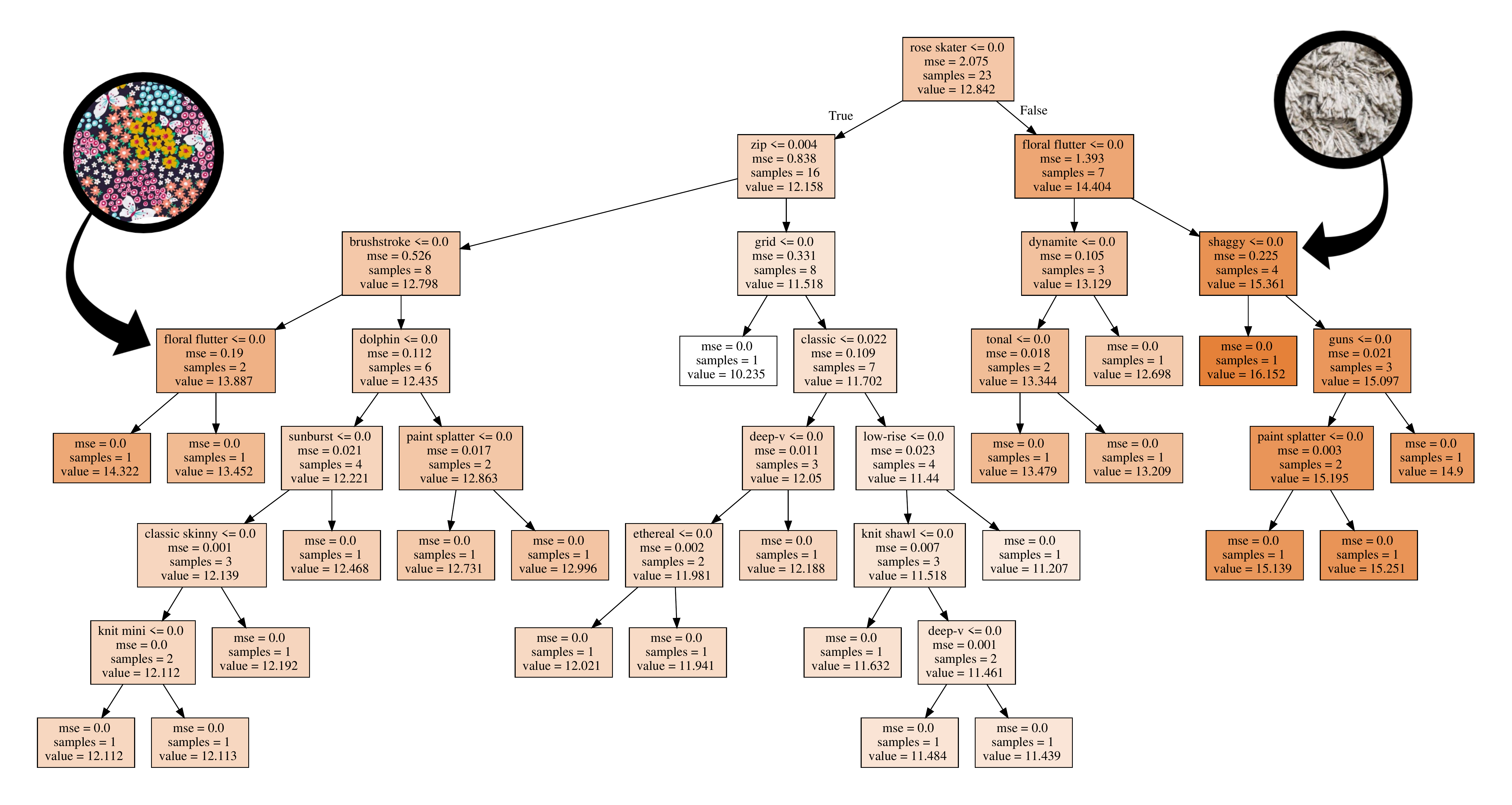}
    \caption{ Feature importance corresponding to a user.}
    \label{fig: treemapping}
\end{figure*}

The performance comparison among different approaches can be seen in Table \ref{table: modelcomparsion}. We can further infer the following from the table: 
\begin{enumerate}
    \item The content-based scheme performs better than a sequential model in modeling the general compatibility since BPR-DAE shows superiority over Bi-LSTM.
    \item Both the modalities, i.e., contextual and visual, are important in modeling the personal preference because VTBPR outperforms VBPR, TBPR and BPR-MF.
    \item GP-BPR achieves better performance than all the other baselines that focus on either the general compatibility modeling or personal preference modeling. This confirms that it is essential to incorporate both the general item-item compatibility and user-item preference in the context of personalized clothing matching. 
    \item Lastly, our model outperforms all other approaches, validating the advantage of a fine-grained semantic attribute representation for a fashion item.

\end{enumerate}

We further evaluate the significance of each modality in our model by comparing PAI-BPR with its two derivatives: PAI-BPR-V and
PAI-BPR-T, where the visual and contextual modality of fashion
items were individually explored, respectively. The performance comparison of these three models can also be seen in Table \ref{table: modelcomparsion}. It can be observed that PAI-BPR outperforms both its derivatives, indicating that the visual and contextual signals complement each other and both contribute to the personalized compatibility modeling. Moreover, a comparison between the textual and visual variants of both VBPR and PAI-BPR shows that the textual variants perform better. This may be due to multiple reasons. Firstly, the key features of fashion items like the pattern and style can be effectively summarized in the contextual information. Additionally, the contextual data usually conveys some high-level semantic cues like the item brand. This captures the brand preference of a user along with modeling the general compatibility, as items of the same brand are more likely to be compatible.

\subsection{Fashion Item Retrieval}
To assess the practical applicability of our model, we evaluate it on personalized and compatible fashion item retrieval. Following the evaluation procedure adopted by Song et al. \cite{song_gp-bpr_2019}, each query consisted of a user-top pair ($u_{m},t_{i}$) which was fed in $D_{test}$, and \textit{T} bottoms were randomly selected as the ranking candidates. Thereafter, these bottoms are ranked for each query on the basis of the compatibility score. The average position of the positive bottom in the ranking list was considered and thus the mean reciprocal rank (MRR) was adopted as a metric \cite{jiang_fast_2015, 7950936, 8215531}. 

\begin{figure*}[!t]
    \centering
    \includegraphics[width=1\linewidth]{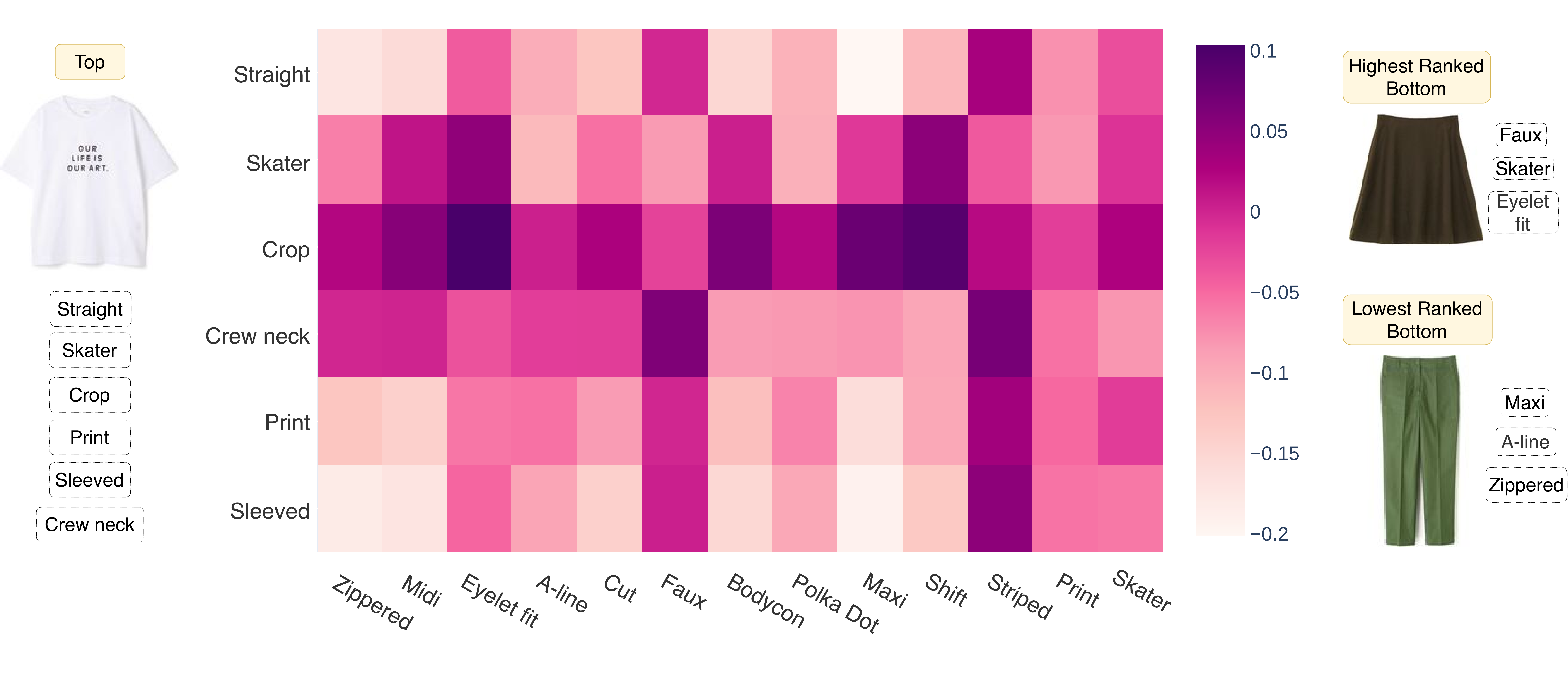}
    \caption{ Pairwise Attribute Correlation among attributes of selected tops and bottoms for a given user.}
    \label{fig: heatmap}
\end{figure*}

Since the dataset is collected from IQON, where the users combine a set of clothing items to make an outfit and share with the world, we consider these outfits as a user's historical preference. Figure \ref{fig: bottomranking} shows the ranking of a selected set of bottoms for the query `white top', with two users having different historical preferences. The figure shows a random set of three top-bottom pairs that the users have made in the past. In order to determine the effectiveness of our model on recommending a bottom from an unseen set of bottoms for an unseen top, the query top and the test bottoms were not a part of any outfit that the users have created in the past. It can be observed that the ranking changes with the change in the user, or more particularly user preferences. For example, the ranking for User 1 was not only influenced by the presence of a similar white top paired with a skirt in her historical preference but also by the user's preference for skirts and dark colours in general. The second most preferred bottom is a pair of jeans that is similar to the jeans which the user paired with a similar black top, as shown in the second user preference of User 1. The rest of the bottoms are also ranked appropriately in accordance with the user's preferences. Similarly, since User 2 had previously paired two white tops with denim bottoms, the top two bottoms retrieved were also denims.  

\subsection{Attribute Preference Identification}
To intuitively understand a user's preference of one bottom over the other for a given top, we identify the most important features that a user looks for in a bottom (or top), followed by a correlation analysis of pairs of attributes. We use a Feature Importance based Architecture like a Decision Tree Regressor with attribute vectors from the Attribute Classification Network as inputs in the form: [$\Delta_{top}$ ,$\Delta_{+ bottom}$, $\Delta_{- bottom}$]  mapped to the output from PAI-BPR so as to understand which attributes/features are important in predicting the output. Figure \ref{fig: treemapping} shows the decision tree generated for a particular user from the Feature Importance Architecture with attributes as the nodes of the tree. Except the leaf nodes, each node contains the name of the attribute, the number of samples in which the attribute was present and the importance value of that attribute. Darker nodes signify higher importance values, i.e. more important features for a particular user. This helps us extract and rank the most preferred attributes in a top and a bottom for a given user.

For analysing the underlying reason behind the ranking of bottoms for a given top as shown in Figure \ref{fig: bottomranking}, we find the correlation between every attribute corresponding to the top and every attribute present in the set of all bottoms. Figure \ref{fig: heatmap} shows the correlation between the 6 most prominent attributes in the top and the 3 most prominent attributes in each bottom. Since some bottoms had overlapping attributes, we had a total of 13 distinct attributes present in the set of bottoms. The bottom attributes are present on the X-axis while Y-axis signifies top's attributes. The correlation between attribute pairs is in line with the ranking results. To find out the overall correlation of a bottom attribute, we calculate the sum of correlation between the bottom attribute and all the attributes in the top. For example, the attributes \textit{Faux} and \textit{Skater}, which are present in the top-ranked bottom for User 1, have the maximum overall correlation with top attributes. Also, even though the third-ranked bottom has a highly correlated attribute \textit{Striped}, its other attributes like \textit{Maxi} and \textit{A-line} contribute to its low rank. Lastly, the two lowest ranked bottoms have the least correlated attributes like \textit{Polka dot}, \textit{Maxi}, \textit{A-line}, \textit{Print} and \textit{Zippered}. To the best of our knowledge, this is the first time such analyses have been done to draw these inferences. These can be helpful in creating much better recommendation systems for fashion-centric portals, which would maximize recommendation hits and subsequently the revenue.

\section{Conclusion and Future Work}
\label{section: conclusion}
In this work, we present an outfit recommendation scheme towards interpretable personalized clothing matching, called PAI-BPR, which models the compatibility between fashion items based on both general aesthetics and personal preferences of a user. We deploy a fine-grained attribute representation model to extract the visual features of an item to overcome the limitation of deep learning techniques in providing explainable results. Apart from visual features, we also incorporate the textual description and categorical information associated with an item since both the modalities are found to deliver valuable information, and together increase the overall performance of the model. In future, we plan to explore multiple ways of multi-modal fusion in order to maximise the information gain. Nevertheless, extensive experiments have been conducted on a real-world dataset and our proposed model outperforms the state-of-the-art, demonstrating the effectiveness of PAI-BPR. Finally, with the help of our attribute representation model, we are successfully able to identify the prominent attribute-pairs in an outfit which make it compatible or incompatible. However, PAI-BPR currently works only for a pair of top and bottom and we would like to extend it to incorporate multiple items like footwear and accessories to comprise an entire outfit in the future.

\bibliographystyle{IEEEtran}
\bibliography{sample-base}

\end{document}